\documentclass[10pt,twocolumn,letterpaper]{article}

\usepackage{wacv}              %

\definecolor{wacvblue}{rgb}{0.21,0.49,0.74}
\usepackage[pagebackref,breaklinks,colorlinks,allcolors=wacvblue]{hyperref}

\def\wacvPaperID{111} %
\def\confName{WACV}
\def\confYear{2026}

\usepackage{times}
\usepackage{epsfig}
\usepackage{graphicx}
\usepackage{amsmath}
\usepackage{amssymb}
\usepackage{booktabs}   %
\usepackage{multirow}   %
\usepackage{array}      %

\usepackage{pifont}
\usepackage{makecell} 
\usepackage{afterpage} %

\newcommand{\cmark}{\ding{51}}
\newcommand{\xmark}{\ding{55}}

\newcommand{\posmine}{PosMine\ }
\newcommand{\clustermine}{ClusterMine\ }

\newcommand{\gt}{\ensuremath\mathcal{Y}_{\text{GT}}}
\newcommand{\yood}{\ensuremath\mathcal{Y}_{\text{OOD}}}
\newcommand{\yreal}{\ensuremath\mathcal{Y}_{\text{real}}}

\newcommand{\neglabels}{\ensuremath\mathcal{Y}_{\text{neg}}}
\newcommand{\yneg}{\ensuremath\mathcal{Y}_{\text{neg}}}
\newcommand{\yaug}{\ensuremath\mathcal{Y}_{\text{aug}}}
\newcommand{\zneg}{\ensuremath\mathcal{Z}_{\text{neg}}}

\newcommand{\poslabels}{\ensuremath\mathcal{Y}_{\text{pos}}}
\newcommand{\ypos}{\ensuremath\mathcal{Y}_{\text{pos}}}
\newcommand{\zpos}{\ensuremath\mathcal{Z}_{\text{pos}}}

\newcommand{\corpus}{\ensuremath\mathcal{Y}_{\text{corpus}}}

\newcommand{\zcorpus}{\ensuremath\mathcal{Z}_{\text{corpus}}}

\usepackage{xcolor}

\usepackage{subcaption}

\title{ClusterMine: Robust Label-Free Visual Out-Of-Distribution Detection via Concept Mining from Text Corpora}

\author{
Nikolas Adaloglou\\
Heinrich Heine University of Düsseldorf \\
{\tt\small adaloglo@hhu.de\thanks{Corresponding author.}}
\and
Diana Petrusheva\thanks{The authors contributed equally. Random order.}\\
Heinrich Heine University of Düsseldorf \\
{\tt\small diana.petrusheva@hhu.de}
\and
Mohamed Asker\footnotemark[2]\\
Heinrich Heine University of Düsseldorf \\
{\tt\small dey59qad@hhu.de  }
\and
Felix Michels\\
Heinrich Heine University of Düsseldorf \\
{\tt\small felix.michels@hhu.de}
\and
Markus Kollmann\\
Heinrich Heine University of Düsseldorf \\
{\tt\small markus.kollmann@hhu.de}}

\begin{document}
\maketitle

\begin{abstract}
Large-scale visual out-of-distribution (OOD) detection has witnessed remarkable progress by leveraging vision-language models such as CLIP. However, a significant limitation of current methods is their reliance on a pre-defined set of in-distribution (ID) ground-truth label names (positives). These fixed label names can be unavailable, unreliable at scale, or become less relevant due to in-distribution shifts after deployment. Towards truly unsupervised OOD detection, we utilize widely available text corpora for positive label mining, bypassing the need for positives. In this paper, we utilize widely available text corpora for positive label mining under a general concept mining paradigm. Within this framework, we propose ClusterMine, a novel positive label mining method. ClusterMine is the first method to achieve state-of-the-art OOD detection performance without access to positive labels. It extracts positive concepts from a large text corpus by combining visual-only sample consistency (via clustering) and zero-shot image-text consistency. Our experimental study reveals that ClusterMine is scalable across a plethora of CLIP models and achieves state-of-the-art robustness to covariate in-distribution shifts. The code is availiable at \url{https://github.com/HHU-MMBS/clustermine_wacv_official}.
\end{abstract}

\let\emptyset\varnothing
\section{Introduction}

Given a set of training images comprising the in-distribution (ID), visual out-of-distribution (OOD) detection aims to identify images sampled from a shifted distribution while having access only to the ID \cite{yang2023imagenet_ood,adaloglou2023adapting}. For each ID, there exists a predefined set of semantic categories $\yreal$ that can be assigned to each sample \cite{zhang2023openood15}. 

Recently, distribution shifts have been categorized into semantic shifts ($\yood \cap \yreal = \emptyset $) and non-semantic shifts, also known as covariate shifts. A covariate-shifted ID pair $(x,y)$ preserves its label $y\in \yreal $ while $x$ undergoes a distribution shift \cite{bai2023feed,zhang2024best}. In the context of natural image recognition, examples of covariate shifts can arise from camera variations, background changes, or style shifts \cite{hendrycks2019oodrobust_in_c,geirhos2018imagenettrained,wang2019learning}.

\textbf{Motivation.} In practice, the ID images exhibit varying levels of class specificity and overlapping concepts \cite{rosch1976basic,hollink2019fruit,miller1995wordnet,adaloglou2024scaling,adaloglou2024rethinking}. While $\gt$ often being a good proxy of $\yreal$ in small data regimes, GT label names at scale can be incomplete, unreliable, or arbitrarily defined. Deriving a comprehensive set of visual concepts that fully characterizes the ID is challenging \cite{vo2024automatic}. For instance, automated curation methods, such as hashtags \cite{singh2022revisiting_hashtag,mahajan2018exploring_hash}, are frequently underdescriptive or inconsistent with natural language \cite{dalle3,xu2024demystifying_metaclip}. Fixed label names can bottleneck performance when class semantics shift due to in-distribution drifts, a common occurrence after deployment. To overcome these challenges, our work proposes to extract ID concepts directly from large text corpora that are aligned with the ID images.

\begin{figure*}
\centering\includegraphics[width=0.96\linewidth]{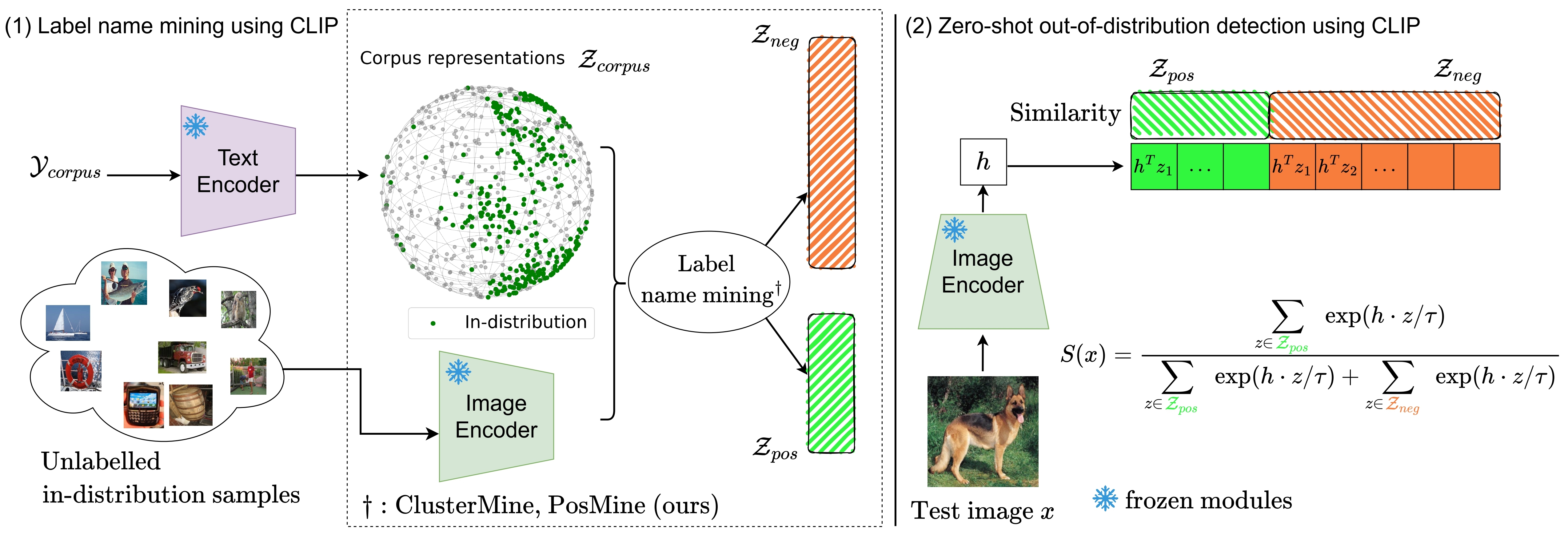}
    \caption{\textbf{An overview of the label mining framework for OOD detection using CLIP}. Given a text corpus $\corpus$ and its feature representation $\zcorpus$, ClusterMine and PosMine aim to extract in-distribution-related class names $\ypos$ in the shared vision-language space of CLIP. $\zneg$ can be either realized as the non-overlapping elements of $\ypos$ and $\corpus$, or as the most dissimilar text representations from $\zpos$ (negative label mining). The OOD detection score is $S(x)$. Best viewed in color.}
    \label{fig:overview}
\end{figure*}

\textbf{Detecting covariate shifts.} An ideal OOD detector should be sensitive to semantic shifts and, at the same time, robust to covariate shifts \cite{yang2022openood}. Misclassified covariate-shifted samples can undermine the generalization capability of an ID classifier. When these samples are flagged as OOD, they render the classifier's predictions unreliable \cite{zhang2023openood15}. Yang et al.\ \cite{yang2023imagenet_ood} have recently demonstrated that existing supervised OOD detection algorithms are susceptible to covariate shifts. To date, the OOD robustness to covariate shifts has not been sufficiently explored \cite{xu2024demystifying_metaclip,fang2023data_dfn,zhai2023sigmoid_sigclip}.

\textbf{Scalability and vision-language (VL) models.} In parallel, VL models have been established for large-scale OOD detection benchmarks \cite{radford2021clip,galil2023a,ming2022mcm,esmaeilpour2022zero,adaloglou2023adapting}. However, the majority of literature comparisons \cite{ming2023does,ming2022mcm,zhang2024ada_neg,wang2023clipn_neg} are conducted with small-sized models relative to the pretraining dataset, such as CLIP ViT-B \cite{vit,radford2021clip}. Smaller-sized models bottleneck performance when other scaling factors remain constant \cite{kaplan2020scaling,cherti2023reproducible_clip_g}. This raises questions whether the recently developed methods are scalable with respect to model size. Plain image-text similarities have shown non-optimal performance as the model size increases \cite{adaloglou2023adapting}. Last but not least, current VL methods for OOD detection \emph{assume access to} $\gt$, which simplifies the task at hand.

\textbf{Questionable benchmarks.} Another point of criticism concerns the adopted large-scale OOD benchmarks. Huang et al. \cite{huang2021mos} introduced a quadruplet of OOD benchmarks for ImageNet with unknown levels of semantic overlap and limited diversity. This benchmark suite remains widely adopted to date \cite{jiang2024negative,ming2022mcm,wang2023clipn_neg}, despite recent studies raising concerns about its validity \cite{yang2023imagenet_ood}. For instance, Bitterwolf et al. \cite{bitterwolf2023ninco} estimate a semantic overlap of almost 60\% in the Places OOD benchmark compared to ImageNet \cite{deng2009imagenet}. Several newer datasets have been carefully designed to minimize semantic overlap with the ID \cite{bitterwolf2023ninco}.

The above factors - \textit{reliance of VL models on ID label names, covariate shifts, scalability, benchmarks} - OOD detection methods may have begun to overfit the idiosyncrasies of certain models and benchmarks, mirroring concurrent trends observed in image classification \cite{beyer2020we}. In other words, it is unclear how robust the recent advancements in large-scale OOD detection using CLIP models are. 

In this paper, we present a general framework for extracting relevant concepts from a large text corpus that encompasses existing methods. Moreover, we develop the first positive label name mining method, ClusterMine, which extracts high-quality ID-related concepts from the corpus. A systematic large-scale OOD benchmarking reveals that ClusterMine i) scales with model size and performance, ii) achieves state-of-the-art OOD detection AUROC on most benchmarks \emph{without access to ground truth ID label names}, and iii) is robust against most ID shifts on ImageNet, iv) outperforms state-of-the-art approaches in near-OOD benchmarks, such as AdaNeg \cite{zhang2024ada_neg} that impose additional requirements (i.e. sequential access to OOD samples).

\section{Related work}
\textbf{Supervised visual OOD detection methods.} Supervised OOD detection involves training or fine-tuning a classifier on labeled ID samples \cite{liang2018enhancing,jiang2023detecting,wang2021can,lee2018mahalanobis}. Posthoc detectors then typically compute an OOD score such as the maximum softmax probability (MSP) \cite{hendrycks2017msp}. Alternative training-time modifications include regularizations by incorporating OOD data \cite{liu2020energy,wang2021energy,liang2018enhancing}, auxiliary samples \cite{hendrycks2018deep_outlier}, or synthetic outliers \cite{du2022towards_vos}. However, when benchmarked rigorously and at scale, Yang et al.\ \cite{yang2023imagenet_ood} demonstrate that no OOD detection score \cite{liu2020energy,hendrycks2022scalingood,zhang2023decoupling,sun2022oodknn,sun2021react} consistently outperforms MSP when applied to a supervised classifier, which has shifted the focus to VL models.

\textbf{OOD detection using CLIP.} Fort et al. \cite{fort2021exploring} utilize CLIP by combining ID and OOD candidate class names for zero-shot inference. Nonetheless, prior knowledge of OOD class names is rarely available in real-world scenarios. Subsequently, Ming et al. \cite{ming2022mcm} show that simply using MSP on image-text similarities, commonly referred to as \emph{maximum concept matching (MCM)}, is sufficient for OOD detection using CLIP. Here, the text representations are computed from the ID label names. In \cite{esmaeilpour2022zero}, the authors introduce an additional text decoder on top of CLIP’s image encoder to generate candidate labels, which can result in overlapping labels with ID data. Using artificially generated OOD data, \cite{tao2023nonparametric_clip,li2025synood_iccv} finetune CLIP to learn a decision boundary between ID and OOD data. 

Recently, Galil et al. \cite{galil2023a} demonstrated that CLIP can function as a capable zero-shot detector without fine-tuning. To date, the only approach that showed promising scaling behavior is by Adaloglou et al. \cite{adaloglou2023adapting}, who propose a two-step approach to train a head using pseudo-labels derived from zero-shot inference. Additionally, CLIP provides a flexible way to define which classes are considered ID during inference \cite{galil2023a}. This aspect remains relatively unexplored, as prior methods rely on the predefined set of ground-truth ID class names, which may not always be optimal.

\textbf{Incorporating negative concepts into CLIP.} A promising direction using CLIP models is the integration of negative prompts or negative labels \cite{fort2021exploring,wang2023clipn_neg,li2024learning_neg,jiang2024negative,zhang2024ada_neg}. Negative prompt learning methods introduce a learnable ``negative'' prompt along with a dedicated ``negative'' text encoder to capture negation semantics in images \cite{wang2023clipn_neg}. Such an approach requires auxiliary data and the training of additional text-based components. Li et al. \cite{li2024learning_neg} circumvent these limitations by learning negative text prompts using CLIP’s existing text encoder, enabling the model to specialize in capturing negative semantics relative to ID classes.

Negative labels refer to adding class names during zero-shot inference. The additional class names likely capture OOD-related classes. A naive approach is to use the $\yood$, as employed by Fort et al.\cite{fort2021exploring}. Our work is more closely related to NegLabel \cite{jiang2024negative}, a training-free method that identifies negative labels based on their dissimilarity from $\gt$ using a text corpus $\corpus$. More recently, \cite{zhang2024ada_neg} developed adaptive variants of NegLabel that dynamically update the negative concepts at test time as more OOD test images become available.

\textbf{OOD detection robustness to covariate shifts.}
Distribution shifts are typically categorized into covariate and semantic shifts \cite{yang2023full}. Covariate shift is usually associated with model calibration \cite{tian2021exploring,guo2017calibration,ovadia2019can_calibration,chan2020unlabelled_calibration}. Yang et al. \cite{yang2023full} coined the term ``full-spectrum detection'' in the context of small-scale benchmarks. Full-spectrum simultaneously considers semantic shifts (OOD detection) and covariate shifts (also known as OOD robustness or OOD generalization \cite{hendrycks2019oodrobust_in_c,hendrycks2021many,tian2021exploring}) within the OOD evaluation pipeline. Modern supervised OOD detection algorithms remain quite susceptible to non-semantic covariate shifts \cite{yang2022openood,zhang2023openood15}. By extracting negative concepts from a text corpus, NegLabel \cite{jiang2024negative} reports notable improvements in robustness to covariate shifts compared to standard MCM. This suggests that CLIP can be tailored to the ID by selecting suitable negative and positive textual inputs. %

\section{OOD detection methods using CLIP}
\textbf{Notation.} We define a sufficiently large text corpus of possible label names $\corpus = \{y_1,y_2, \dots, y_N \}$ with cardinality $|\corpus|=N$. By mining ID images, we extract from the corpus positive and negative label sets, $\poslabels, \neglabels$, with $\poslabels \cap \neglabels=\emptyset$ and $\poslabels \cup \neglabels \subseteq \corpus$. The set $\corpus$ is designed such that $\poslabels$ is strongly related to the real semantic categories of the ID, whereas $\neglabels$ is not. We use the text encoder of CLIP to generate the representations $\mathcal{Z}_{pos}, \mathcal{Z}_{neg}$, which correspond to $\poslabels, \neglabels$, respectively. We demonstrate the general framework in \cref{fig:overview}, which is described below.

\vspace{-0.1cm}
\subsection{A general OOD detection framework for CLIP} \label{subsec:neglabelmine}
Vision-language models can leverage a secondary set of ``negative'' concepts $\yneg$ that are unrelated to the categories of the ID. Given an image $x$ and its representation $h=g(x)$ from the CLIP image encoder $g(.)$, we can define an OOD score by

\begin{equation}
    S(x) =   \frac{\displaystyle\sum_{z \in \mathcal{Z}_{pos}} \exp(h \cdot z/\tau)}
    {\displaystyle\sum_{z \in \mathcal{Z}_{pos}} \exp(h \cdot z/\tau) + \displaystyle\sum_{z \in \mathcal{Z}_{neg}} \exp(h \cdot z/\tau)},
\label{eq:posneg_score}
\end{equation}

with temperature parameter $\tau>0$. Both image and text representations are unit vectors. The particular choice of $\ypos,\yneg$ depends on the method. For instance, Fort et al. \cite{fort2021exploring} use \cref{eq:posneg_score} with $\ypos=\gt$ and $\yneg=\yood$, which greatly simplifies the task. NegLabel \cite{jiang2024negative} also assumes $\ypos=\gt$ and extracts $\yneg$ from a text corpus $\corpus$ using negative label mining.

\begin{figure}
\centering\includegraphics[width=0.94\linewidth]{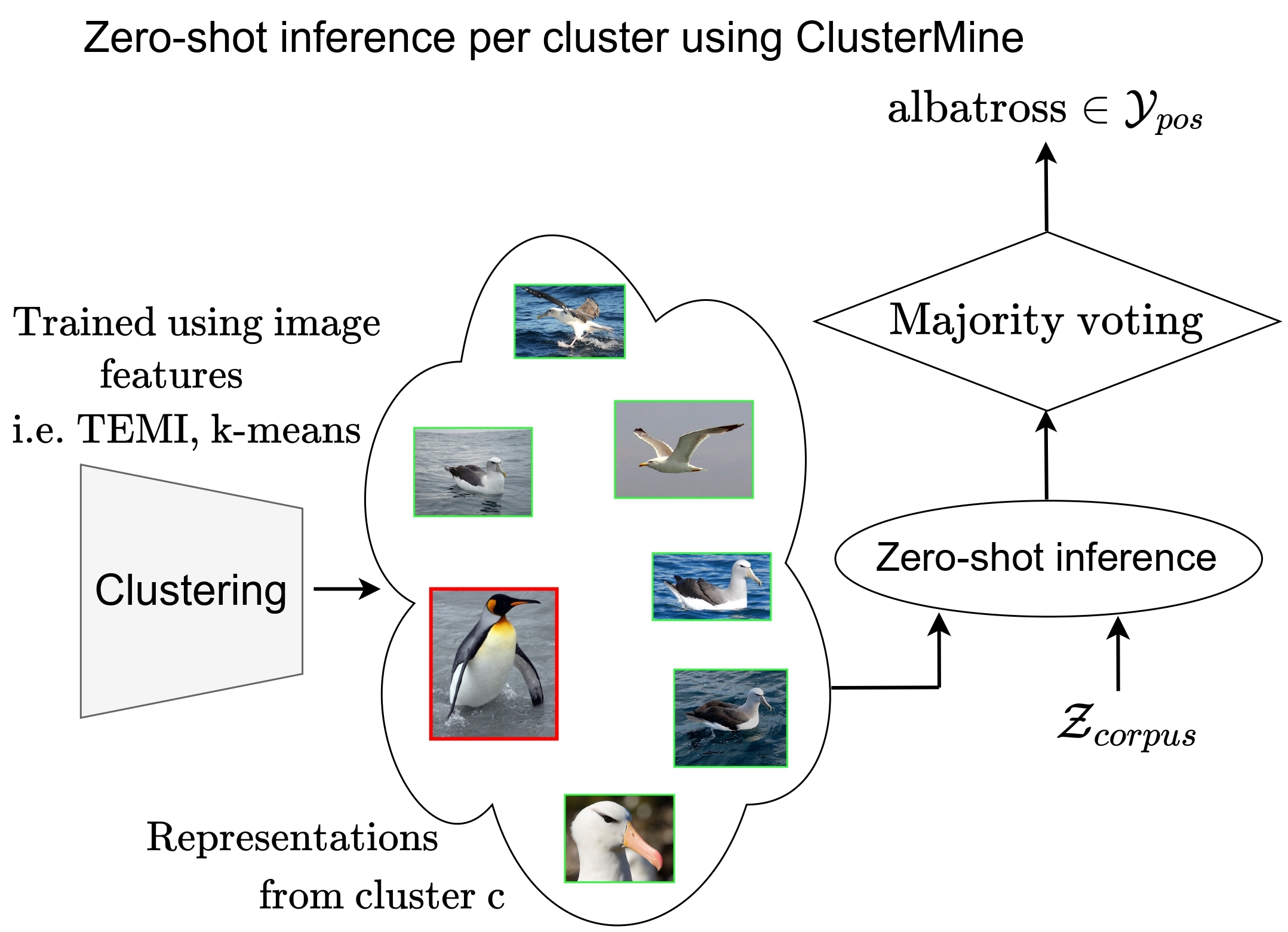}
    \caption{\textbf{A visual illustration of ClusterMine.}}
    \label{fig:clustermine_fig}
    \vspace{-0.5cm}
\end{figure}

\textbf{Negative label mining} \cite{jiang2024negative} first computes the cosine similarities between $\mathcal{Z}_{corpus}$ and $\mathcal{Z}_{pos}$. Excluding overlapping concepts from the corpus, 

\begin{equation}
\yneg = \corpus \setminus  \ypos,
\label{eq:negative_all}  
\end{equation}

the $K\leq|\yneg|$ most dissimilar text representations from $\yneg$ are considered. In \cite{jiang2024negative}, the authors use the percentile distance, i.e. a $95\%$ percentile instead of the minimum. For the edge case of $K=|\neglabels|$, no pruning is applied. 

\textbf{Negative grouping and dynamic negative mining.} By splitting $\neglabels$ into randomly sampled groups, \Cref{eq:posneg_score} can be applied per group \cite{jiang2024negative}. The final score is the average across groups. We did not observe any performance gains from using this strategy (see supplementary material); therefore, it is not employed in this work. When OOD samples appear sequentially, the choice of negative concepts can be adjusted at test time \cite{zhang2024ada_neg}. Such strategies can be integrated into the presented framework, and we leave them for future work. %

\subsection{Positive label mining methods}
Different from previous works \cite{fort2021exploring,jiang2024negative}, positive label mining aims to design OOD approaches using CLIP models that do not depend on prior knowledge of $\gt$. At test time, \cref{eq:posneg_score} is applied after estimating $\ypos, \yneg$ from $\corpus$. Other post-hoc training scores, such as pseudo-label probing \cite{adaloglou2023adapting}, achieved inferior performance (see supplementary material).

We first test whether ``naive'' positive mining from $\corpus$ using zero-shot inference is comparable to existing methods that rely on $\gt$. Afterwards, we consider a label name as ID ($y\in \ypos$) if at least $M$ training samples are assigned to this particular class and $\yneg=\corpus \setminus \ypos$. Alternative formulations of \posmine that explicitly control the cardinality of $\ypos$ can be realized. While results are affirmative (\cref{tab:bigG_results}), selecting $M$ becomes challenging without an OOD validation set. In addition, \posmine solely checks for image-text similarities to derive $\ypos$.

\begin{table*} 
\centering
\resizebox{\textwidth}{!}{%
\begin{tabular}{lccccccc} 
\toprule
\multirow{2}{*}{{Method}}  & \multirow{2}{*}{{NINCO}} & \multirow{2}{*}{{IN-O}}   & \multirow{2}{*}{{OpenImage-O}} & \multirow{2}{*}{{iNat}} & \multirow{2}{*}{{IN-OOD}} &   \multirow{2}{*}{{Textures43}}  & Average     \\ 
&  &&&&&& AUROC/FPR95   \\ 
\midrule
  \multicolumn{6}{l}{\textit{Methods  requiring in-distribution label names $\gt$} for zero-shot inference}\\
    {MaxLogit} \cite{hendrycks2022scalingood} &  83.98 / 58.03 & 90.69 / 39.85 & 92.52 / 35.54 & 92.23 / 41.80 & 90.72 / 44.01 & 87.77 / 51.81 & 89.65 / 45.17 \\
     {MD} \cite{lee2018mahalanobis,ren2021simple} &  85.22 / 64.99 & 91.37 / 43.65 & 91.01 / 59.68 & 87.20 / 90.30 & 90.72 / 50.45 & 94.76 / 31.14 & 90.05 / 56.70 \\
     {Relative MD} &  89.25 / 49.49 & 92.18 / 35.50 & 94.32 / 29.08 & 91.58 / 52.79 & 89.53 / 45.21 & 90.61 / 43.40 & 91.25 / 42.58 \\
  {MCM} \cite{ming2022mcm} & 88.78 / 49.00 & 91.30 / 41.20 & 96.64 / 16.80 & 96.62 / 16.15 & 89.65 / 47.36 & 91.75 / 40.84 & 92.46 / 35.22 \\
  {PLP} \cite{adaloglou2023adapting} &  91.80 / 42.57 & 93.30 / 33.05 &\textbf{97.87} / \textbf{10.84} & 98.94 / 3.86 & 90.01 / 47.42 & \textbf{94.79} / \textbf{26.01} & 94.45 / 27.29 \\
  \midrule   
 \multicolumn{6}{l}{ {\textit{Negative label mining from $\corpus$ given $\gt$}}}\\
 {NegLabel} & 88.7 / 50.12 & 85.29 / 55.7 & 94.61 / 26.08 & 99.4 / \textbf{2.46} & 82.72 / 62.62 & 85.06 / 56.19 & 89.3 / 42.19 \\

 {NegLabel$^*$} &  90.26 / 47.07 & 90.10 / 43.45 & 95.79 / 22.31 & 98.64 / 6.75 & 88.4 / 48.34 & 90.44 / 43.95 & 92.27 / 35.31 \\
 {CLIPScope} &  92.69 / 39.47 & 88.18 / 45.15 & 96.24 / 17.66 & \textbf{99.45} / 2.57 & 91.04 / 43.15 & 90.84 / 43.77 & 93.07 / 31.96 \\
  \midrule   
 \multicolumn{6}{l}{ {\textit{Ours: Positive label mining from $\corpus$}}}\\
 {PosMine}  &  92.56 / 33.53 & 93.13 / 32.3 & 97.04 / 15.8 & 98.83 / 5.83 & 91.36 / 38.89 & 93.81 / 32.1 & 94.46 / 26.41 \\
 {ClusterMine}     &   \textbf{92.87} / \textbf{30.3} & \textbf{93.57} /\textbf{ 29.4} & 96.93 / 15.9 & 99.0 / 4.77 & \textbf{91.53} / \textbf{38.26} & 93.45 / 32.89 & \textbf{94.56 / 25.26} \\
\bottomrule
\end{tabular}} 
\caption{\textbf{Semantic large-scale OOD detection AUROCs ($\uparrow$) / FPR95($\downarrow$) per dataset using CLIP ViT-H dfn5b \cite{fang2023data_dfn}}. The WordNet \cite{miller1995wordnet} corpus is used (nouns and adjectives), and the ID is ImageNet. All reported baselines are reproduced results and do not require training or fine-tuning of CLIP. MD stands for Mahalanobis distance \cite{lee2018mahalanobis}. The best scores are shown in \textbf{bold}. The symbol $*$ indicates tuning $|\yneg|$ to 40000 for NegLabel \cite{jiang2024negative}, which is different from the authors' choice of 10000.} 
\label{tab:bigG_results}
\end{table*}

\subsubsection{Cluster-based positive mining (\clustermine)} 
The proposed method, cluster-based positive mining (\clustermine, \cref{fig:clustermine_fig}), consists of the following steps:
\begin{enumerate}
    \item \textbf{Visual feature-based clustering:} We perform  clustering on the visual encoder of CLIP using $C$ clusters. In practice, we apply TEMI clustering \cite{adaloglou2023exploring} as it has shown significant improvements in clustering accuracy over $k$-means \cite{lloyd1982kmeans}, even at large scales \cite{adaloglou2024scaling}. We use the default parameters ($\beta=0.6$, 50 heads) as in \cite{adaloglou2023exploring}. In contrast to the clustering downstream task, we are only interested in a rough overestimation of $C$.
    \item \textbf{Vision-language inference:} For all samples that fall into the same cluster, we apply zero-shot inference using the text corpus $\corpus$. 
    \item \textbf{Cluster Voting:} Each cluster's label name is then determined by applying majority voting, effectively reducing the false positive classes. The latter enforces visual consistency, as the nearest neighbors in feature space likely share the same label \cite{scan, temi}. Crucially, because different clusters can be mapped to the same label name, $|\ypos| \leq C$. A heuristic for setting $C$ is the ``elbow'' method \cite{marutho2018determination} using the saturation of the ratio $|\ypos|/C$ or the percentage of $|\ypos|$ that appear on multiple clusters as the primary metric, analogous to  \cite{marutho2018determination,adaloglou2024rethinking} (see supplementary material).
\end{enumerate}

\noindent\textbf{Benefits of ClusterMine}. ClusterMine has the following advantages over PosMine. It additionally accounts for image-image similarities within the ID features via clustering, similar to a human. By integrating visual consistency into the top-1 image-text concept from $\corpus$: 1) different clusters can be mapped to the same label name $y\in \ypos$, and 2) text concepts that do not match the samples' neighborhood are rejected (\cref{fig:clustermine_fig}). Thus, $|\ypos|$ becomes relatively insensitive as $C$ increases far above $\yreal$ (\cref{fig:ablations_all}, right), unlike $M$ in \posmine (\cref{fig:ablations_all}, center). In practice, an overestimation of the real semantic categories for selecting $C$ is possible even with minimal to no domain knowledge \cite{scan,adaloglou2024rethinking}. Experimental results show a superior label quality for ClusterMine (\cref{fig:overlap},\cref{tab:corpora_ablation}). Compared to MCM, \clustermine leverages a corpus to decide which concepts are positive and negative. Compared to NegLabel \cite{jiang2024negative}, \clustermine extracts the positives and \emph{implicitly} defines the negative concepts without a pre-defined explicit threshold of NegLabel, which is hard to determine a priori.

\noindent\textbf{Combining positive and negative label mining.} \posmine and \clustermine can be easily combined with the negative mining strategy from \Cref{subsec:neglabelmine}. After computing $\yneg$ using \Cref{eq:negative_all}, the $K$ most dissimilar text representations from $\ypos$ are calculated. Under this prism, previous methods \cite{fort2021exploring,jiang2024negative} can be viewed as special cases of our label mining framework (\Cref{fig:overview}). Interestingly, we show that state-of-the-art large-scale CLIP models do not require pruning $\yneg$ when $\ypos$ are extracted from the corpus. Unless otherwise specified, \posmine and \clustermine default to using $\yneg$ as shown in \Cref{eq:negative_all}.

\section{Experimental evaluation}
\subsection{Training-free OOD detection baselines}
We focus on approaches that do not require training the image or text encoder of CLIP or additional text encoders \cite{wang2023clipn_neg}. To this end, we consider the following CLIP-based OOD detection methods using $\gt$ as baselines: Energy \cite{liu2020energy}, MaxLogit \cite{hendrycks2022scalingood}, Mahalanobis distance (MD), relative MD \cite{lee2018mahalanobis,ren2021rmd}, MCM \cite{ming2022mcm}, and PLP \cite{adaloglou2023adapting}. For instance, MCM defines $S(x)$ as the maximum softmax probability (MSP) of image-text similarities. PLP \cite{adaloglou2023adapting} uses the same pipeline as in MCM, but derives a pseudo-label for each image and trains a linear classifier using pseudo-labels. %

We adopt NegLabel \cite{jiang2024negative} as our primary baseline method, which utilizes an external corpus. We highlight that only \clustermine and \posmine do not assume access to $\gt$. We use $K=4\cdot 10^4$ negative concepts instead of $K=10^4$ used in \cite{jiang2024negative}, which we denote as NegLabel$^*$. Finally, we show a small-scale comparison using ViT-B with recent state-of-the-art methods, such as AdaNeg and SynOOD \cite{zhang2024ada_neg,li2025synood_iccv}, which use additional information or data.

\begin{figure*}[hbtp]
    \centering \includegraphics[width=1\linewidth]{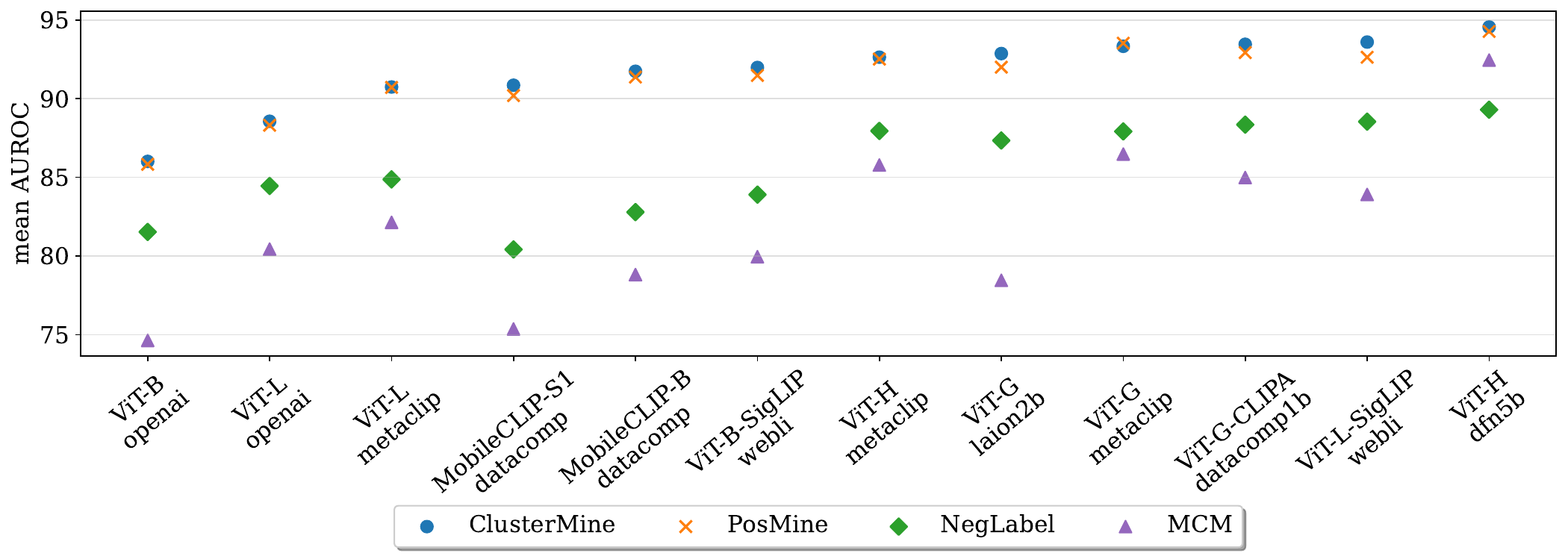} \caption{\textbf{Scalable out-of-distribution detection AUROC (\%, y-axis) using various pretrained CLIP weights (x-axis)}. Unlike previous state-of-the-art methods that require $\gt$ (MCM, NegLabel), \emph{ClusterMine and PosMine} extract the in-distribution-related label names from a text corpus. Mean AUROC is computed across six OOD datasets, using ImageNet-1K as ID. We use the WordNet \cite{miller1995wordnet} corpus. Pretrained CLIP models are sorted based on their performance with respect to ClusterMine.} \label{fig:scaling}
\end{figure*}

\subsection{Implementation details}
\textbf{Visual OOD datasets.} Using ImageNet as ID, we carefully choose six OOD datasets with a sufficiently small degree of semantic overlap based on previous literature. Specifically, we include NINCO \cite{bitterwolf2023ninco}, IN-O \cite{hendrycks2021natural}, OpenImage-O \cite{wang2022vim}, a subset of plant images from iNaturalist (iNat) \cite{huang2021mos}, IN-OOD \cite{yang2023imagenet_ood}, and a subset of Textures called Textures43 \cite{wang2022vim}. IN-OOD and IN-O are constructed from ImageNet-21K, excluding ImageNet samples. NINCO and IN-OOD have a lower degree of semantic overlap, due to their careful manual sample selection. Unless otherwise specified, the mean AUROC and FPR95 are computed from these six out-distributions.

\textbf{Covariate shifted data.} We consider five commonly used covariate-shifted ID in generalization benchmarks, namely ImageNetV2 \cite{recht2019imagenet}, ImageNet-C \cite{hendrycks2019oodrobust_in_c}, ImageNet-A \cite{hendrycks2021nae}, sketches \cite{wang2019learning}, ImageNet-R \cite{hendrycks2021many}. ImageNet-R consists of 30000 samples from 200 ImageNet classes, where images contain stylistic and rendition variations, including cartoons, graphics, sketches, and tattoos. 
ImageNetV2 consists of independently collected samples from Flickr. ImageNet-A is collected using adversarial filtration from supervised classifiers, while ImageNet-C applies 15 pixel-level manipulations, such as adding blurring or weather-like effects. We use a randomly selected subset of 10000 images of ImageNet-C.

\textbf{Text corpora.} We utilize WordNet \cite{miller1995wordnet} as a representative large-scale text corpus. We use all nouns and adjectives by default, resulting in $|\corpus|\approx79 \times 10^3$ concepts. We
also adopt the preprocessed ImageNet-21K (IN21K) \cite{ridnik2021imagenet21k} subset. As a larger-scale corpus, we include Part-of-Speech (POS) Taggings \cite{pos_corpus}, which contains $270\times 10^3$ concepts. We create subsets by considering only nouns (N) or nouns and adjectives (NA). We pre-process the corpus by removing duplicates and using one lemma of the SynSet in Wordnet \cite{miller1995wordnet}. Other choices for text pre-processing had similar results (see supplementary).

\textbf{Vision language models, runtime and memory usage.} We use CLIP ViT-H dfn5b \cite{fang2023data_dfn} weights by default. For the cross-model analysis, we use 12 publicly available weights from \cite{radford2021clip,zhai2023sigmoid_sigclip,openclip,cherti2023reproducible_clip_g, fang2023data_dfn}. We set $\tau$ to be 10 times smaller than the pretraining temperature used in each model. In our experiments, we set $C=4000$ clusters and $M=100$ and present their sensitivity in \cref{fig:ablations_all}. For clustering, we use TEMI \cite{adaloglou2023exploring} with the default parameters. \clustermine remains training-free w.r.t the CLIP parameters, so image and text features can be precomputed. With the largest scale model ViT-G, clustering on ImageNet takes less than 1.5 hours using TEMI on a single GPU with less than 10 GB of VRAM using the default settings. Since text and image representations are pre-computed, inference for all six OOD detection benchmarks can be achieved in a \textit{maximum} of twelve minutes with less than 16GB of VRAM for a mini-batch size of 4096 for ViT-G with the POS corpus.

\begin{figure}
\centering\includegraphics[width=0.96\linewidth]{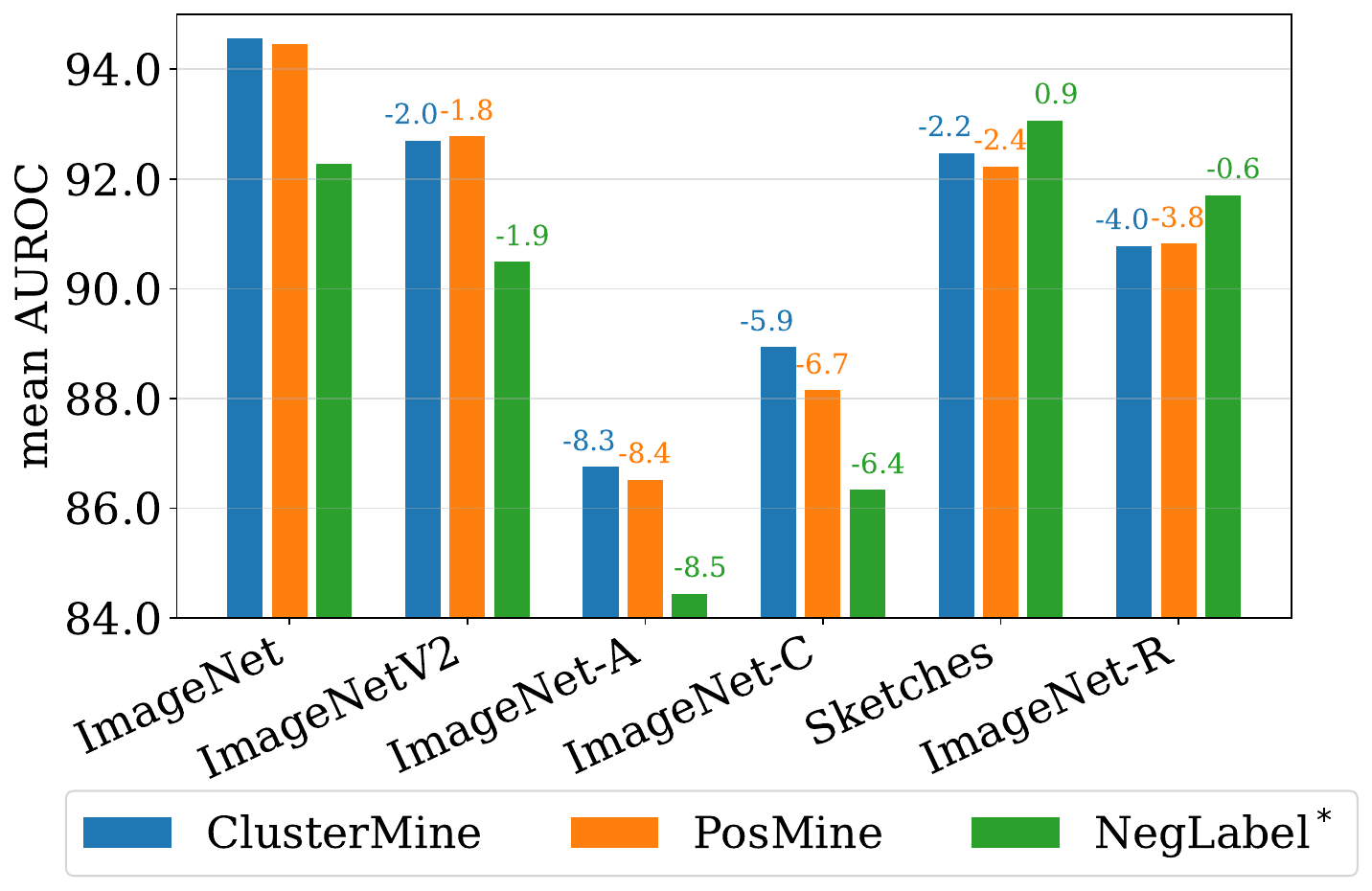}
    \caption{\textbf{OOD detection robustness to multiple ID shifts (x-axis) compared to ImageNet using CLIP ViT-H dfn5b \cite{fang2023data_dfn}}. The relative AUROC difference in \% of each method compared to its ImageNet score is shown on top of each bar. We report the mean AUROC ($\uparrow$,\%) across six different OOD datasets (y-axis).}
    \label{fig:robustness}
\end{figure}

\subsection{Main experimental results}
\textbf{Main results.} \clustermine is the first method the achieves state-of-the-art OOD detection on most ImageNet benchmarks without $\gt$ (\Cref{tab:bigG_results}). Interestingly, \clustermine outperforms in the most curated benchmarks that have the least amount of semantic overlap, namely NINCO, IN-O, and IN-OOD. This reveals the ability of large-scale vision language models to \emph{dynamically} extract positive concepts, given a sufficiently descriptive corpus. This is crucial when in-distribution drifts occur naturally \cite{oquab2023dinov2,vo2024automatic}.

\begin{table}
\centering
\resizebox{0.9\columnwidth}{!}{%
\begin{tabular}{llcc}
\toprule
\multirow{2}{*}{{ID Dataset}} & \multirow{2}{*}{{Method}} & Average & Average     \\ 
  &  & AUROC & FPR95  \\
\hline
  & NegLabel$^*$ &  {\textbf{93.06}} &  {40.29} \\
  & PosMine &  {92.22} &  {32.92} \\
 \multirow{-3}{*}{\makecell{ImageNet-\\Sketches}}  & ClusterMine &  {92.47} &  {\textbf{30.94}} \\
 \midrule
  & NegLabel$^*$ &  {\textbf{91.70}} &  {41.42} \\
  & PosMine &  {90.82} &  {\textbf{35.92}} \\
 \multirow{-3}{*}{ImageNet-R} & ClusterMine &  {90.78} &  {37.50} \\
 \midrule
  & NegLabel$^*$  & 84.43 & 56.32 \\
  & PosMine & 86.52 & 44.44 \\
  \multirow{-3}{*}{ImageNet-A}  & ClusterMine & \textbf{86.76} & \textbf{42.57} \\
  \midrule
  & NegLabel$^*$ & 90.49 & 45.38 \\
  & PosMine &  \textbf{92.78} & 32.15 \\
 \multirow{-3}{*}{ImageNetV2}  & ClusterMine &  92.70 & \textbf{31.87} \\ \midrule
  & NegLabel$^*$ & 86.33 & 54.87 \\
  & PosMine & 88.15 & 52.10 \\
\multirow{-3}{*}{ImageNet-C}  & ClusterMine &  \textbf{88.94} & \textbf{49.21} \\ 
\bottomrule
\end{tabular} }
\label{tab:mini-robust}
\caption{\textbf{OOD detection robustness to covariate shifts}. The star symbol indicates tuning $|\yneg|$ to 40K for NegLabel. The full table for all out-distributions is available in the appendix.}
\end{table}

\begin{table*}
\centering
\resizebox{1.85\columnwidth}{!}{%
\begin{tabular}{lcccccccc}
\toprule
\multirow{2}{*}{{Corpus}} & \multirow{2}{*}{{$|\corpus|$}}  & NegLabel$^*$ &  \multicolumn{3}{c}{{PosMine}}  & \multicolumn{3}{c}{{ClusterMine}}  \\
 \cmidrule(lr){3-3}  \cmidrule(lr){4-6}   \cmidrule(lr){7-9}
& & AUROC & AUROC & Overlap & F1 score & AUROC & Overlap & F1 score  \\
\midrule
  \multicolumn{5}{l}{\textit{WordNet subsets}}\\
    IN21K & 20K & 92.38 & \textbf{95.12}& 93.5& 57.9& 94.66 &83.9&71.1\\ %
    N  & 67K & 92.31 & 94.39 & 87.7& 52.4&\textbf{94.44} &77.2&61.4 \\  %
    NA  & 79K & 92.27 & 94.46 & 87.1& 51.9&\textbf{94.56} & 76.0&60.0 \\ %
\hline
  \multicolumn{5}{l}{\textit{POS subsets}}\\

N/N$\cup\gt$& 231K & {91.75}/91.26 & {92.79}/93.95 & 33.2/78.8 & 19.3/46.6 &\textbf{{92.94}}/\textbf{94.14}& 27.8/69.5 & 21.3/53.5 \\
NA/NA$\cup\gt$ & $\approx$270K & {91.78}/91.27 & {92.69}/94.01 & 33.8/79.1 & 19.7/46.9& \textbf{{92.90}}/ \textbf{94.10}  & 28.4/70.0 & 21.6/53.7\\
\bottomrule
\end{tabular}}
\caption{\textbf{Average OOD detection AUROC across 6 OOD datasets using various text corpora and CLIP VIT-H dfn5b \cite{fang2023data_dfn}}. The subsets ``N'' refer to nouns only, and NA refers to nouns and adjectives. For the POS corpora, we show the impact of manually adding the missing $\gt$ ($\cup\gt$). The class overlap (in \%) is defined as $|\gt \cap \ypos|/ |\gt|$ and the F1 score (in \%) is measured between $\ypos$ and $\gt$.}
\label{tab:corpora_ablation}
\end{table*}

\textbf{Scalability.} Here, we investigate if positive label mining methods remain competitive across various scales of pretrained networks (\Cref{fig:scaling}). We refer to scale as the combination of computing resources, model size, and pretraining data. Apart from model size (i.e., number of parameters), existing models differ in terms of compute time and training data. For this reason, we rank the models based on their OOD detection using \clustermine in \Cref{fig:scaling}. We observe that \clustermine consistently outperforms NegLabel and MCM across 12 model weights by sizable margins in AUROC. Our results suggest that the ability of CLIP to capture ID-related concepts does not depend on the scale.

\textbf{Covariate ID shifts and corpus sensitivity.} By varying the ID while keeping the same OOD data, we measure the average AUROC degradation, as shown in \cref{fig:robustness}. Both \clustermine outperform NegLabel on 4 out of 6 IDs. Specifically for the image-level corrupted version of ImageNet (ImageNet-C), we find that \clustermine is the most robust method for pixel-level perturbations. This suggests that including $\yneg$ using \Cref{eq:posneg_score} is likely the primary driver of OOD detection robustness, similar to proximal works \cite{fort2021exploring,wang2023clipn_neg,li2024learning_neg,jiang2024negative,zhang2024ada_neg}.On Sketches and ImageNet-R, NegLabel seems slightly more robust as measured by AUROC when stylistic changes occur in the ID in \cref{fig:robustness}, whereas \clustermine is superior when comparing FPR95, as shown in \Cref{tab:mini-robust}. While stylistic concepts coexist in the WordNet corpus, we find that the superior AUROC of NegLabel on ImageNet-R and Sketches is not attributed to the negative label mining (see supplementary material).

As illustrated in \cref{tab:corpora_ablation}, \clustermine shows superior performance in all but one corpus. Even when using the complete POS corpus with 270K concepts, we report competitive performance with PLP \cite{adaloglou2023adapting}, the best-performing method using $\gt$. Manually adding the missing $\gt$ labels on the POS corpora improves the OOD detection AUROC for \clustermine, suggesting that domain-specific concepts are beneficial. Although this might be a limitation for novel application domains, it is not necessary to outperform NegLabel in \cref{tab:corpora_ablation}.

\begin{table}[h]
    \centering
    \begin{tabular}{ccc}
    \toprule
     $|\ypos|$:    & $1000$ & $1500$ \\
         \hline
      MCM ($\gt\cup \yaug $)  &  92.46/35.22 & 92.01/36.17 \\ 
     Eq.1 ($\gt\cup \yaug$ )    & 93.08/32.00 & 92.88/32.81 \\
      ClusterMine (C=1.2K, 4K)   & \textbf{93.84/30.74} & \textbf{94.56/25.26} \\
      \bottomrule
    \end{tabular}%
    \caption{\textbf{Average AUROC/FPR95 for equal $|\ypos|$.} In the first two rows, additional positive label names are mined from $\gt$}.
    \label{tab:results_controlled}  
\end{table}

\begin{table*}[h]
    \centering
    \resizebox{\textwidth}{!}{%
    \begin{tabular}{lcccc|cccccc|cc}
        \toprule
        & \multicolumn{4}{c}{Near OOD} & \multicolumn{6}{c}{Far OOD} & \multicolumn{2}{c}{Mean} \\
         \cmidrule(lr){2-7} \cmidrule(lr){8-11} \cmidrule(lr){12-13}
        Method  & \multicolumn{2}{c}{SSB-hard} & \multicolumn{2}{c}{NINCO} & \multicolumn{2}{c}{iNat} & \multicolumn{2}{c}{Textures} & \multicolumn{2}{c}{OpenImage-O} & \multicolumn{2}{c}{(Near/Far)} \\
        \cmidrule(lr){2-3} \cmidrule(lr){4-5} \cmidrule(lr){6-7} \cmidrule(lr){8-9} \cmidrule(lr){10-11} \cmidrule(lr){12-13}
        & FPR95 & AUROC & FPR95 & AUROC & FPR95 & AUROC & FPR95 & AUROC & FPR95 & AUROC & FPR95 & AUROC \\
        \midrule
         {AdaNeg}  & 74.91 & 75.11 & 60.10 & 78.30 & \textbf{0.72} & \textbf{99.72} & \textbf{21.40} & \textbf{95.71} & \textbf{29.81} & \textbf{93.87} & 67.5/17.3 & 76.7/96.4   \\
         {SynOOD}  &-&-&-&-& 1.57 &99.57 & 22.94 & 95.29 &-&-& 71.7/17.1 & 77.6/96.2  \\
        \midrule
         {ClusterMine}  & \textbf{68.51} & \textbf{76.33} & \textbf{54.01} & \textbf{84.16} & 18.01 & 96.30 & 67.69 & 79.97 & 30.72 & 92.55 & \textbf{61.3}/38.8 & \textbf{80.3}/89.6     \\
        \bottomrule
    \end{tabular}
    }
    \caption{ {\textbf{ClusterMine comparison using the CLIP ViT-B weights \cite{radford2021clip}}. We evaluate using the OpenOOD Near- and Far-OOD benchmarks compared to recent state-of-the-art methods. Results for AdaNeg and SynOOD are taken from the reported results, when available.}}\label{tab:new_sota_vitb}
    \vspace{-0.1cm}
\end{table*}

\subsubsection{Mined label quality and controlled experiments}
\begin{figure}
\centering
\includegraphics[width=0.94\linewidth]{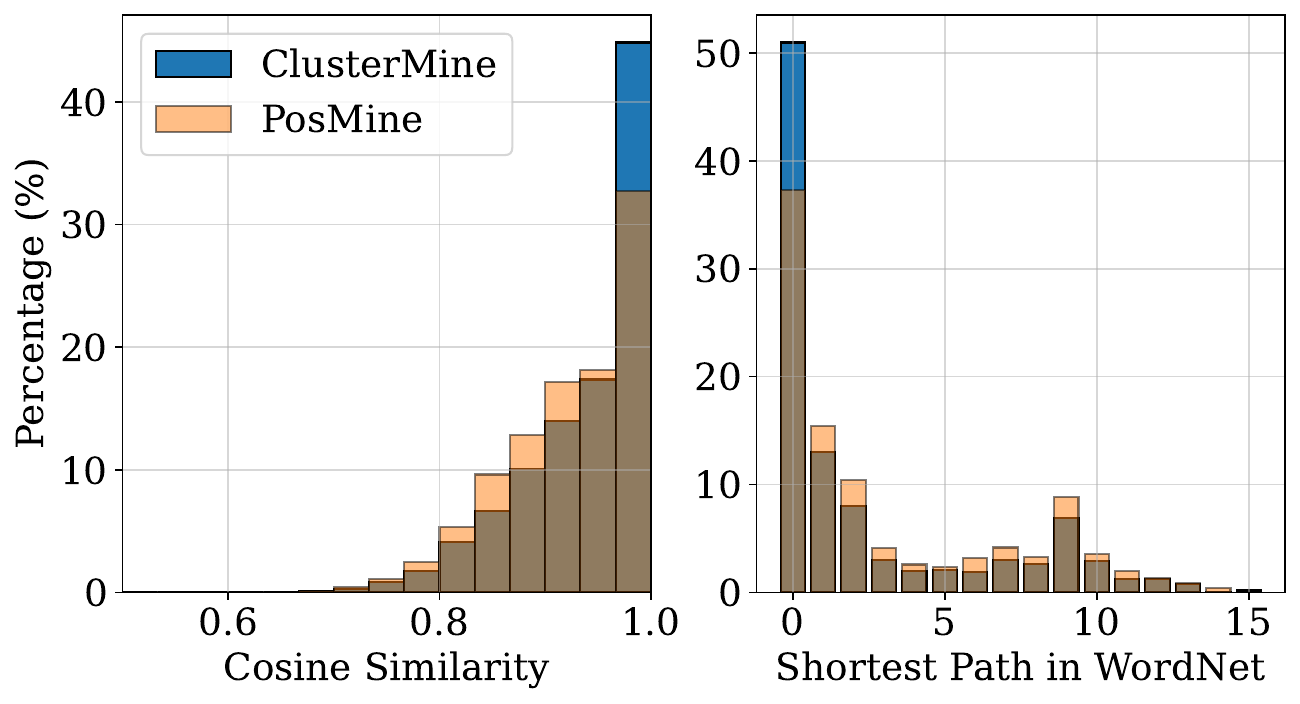}
\caption{\textbf{Analysis of mined label name quality.} We calculate top-1 text-text similarity with GT (left), and by finding the shortest path (minimum amount of hops) to GT in WordNet (right).}
\vspace{-0.65cm}
\label{fig:overlap}
\end{figure}

Since $|\ypos|$ is implicitly determined, we find the F1 score between $\ypos$ and $\gt$ a more suitable comparison compared to the percentage of overlapping classes ($|\gt \cap \ypos|/ |\gt|$). Interestingly, \clustermine achieves a superior F1 score for all text corpora (\cref{tab:corpora_ablation}). This is primarily attributed to the cluster voting strategy, leading to \emph{fewer false positives}. In addition, we visualize the semantic alignment of $\ypos$ with $\gt$ using two normalized histograms. In \cref{fig:overlap} left, we measure the top-1 cosine similarity using CLIP text representations to $\gt$, and in \cref{fig:overlap} right, we measure the shortest path (number of hops) in WordNet. Additionally, intercluster purity remains high ($>50\%$), while the percentage of mined $|\ypos|$ that appear across multiple clusters increases as $C$ increases, confirming that ClusterMine maintains semantic consistency while being robust to the overestimation of $C$ (see supplementary material).

\begin{figure*}
\centering\includegraphics[width=0.95\linewidth]{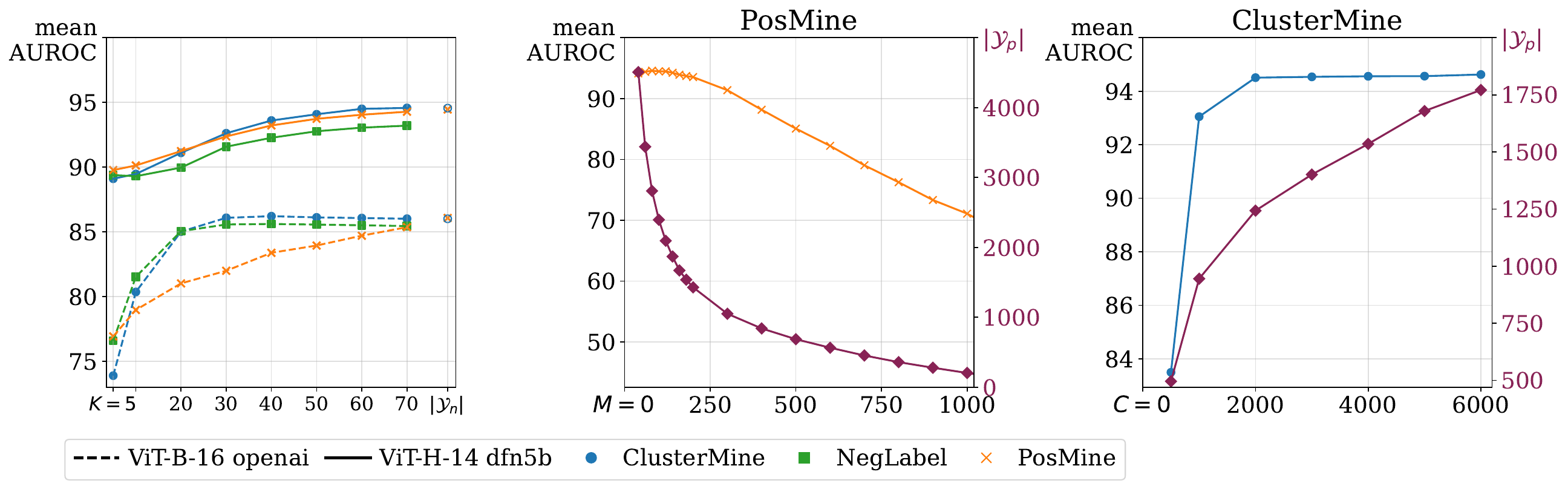}
\caption{\textbf{Sensitivity to hyperparameters for negative label mining for all methods (left), \posmine (center), and \clustermine (right).} We vary the number of negative labels $K$ (in thousands, \textbf{left}), minimum number of assigned samples per class $M$ using \posmine (\textbf{center}), and number of clusters used in \clustermine (\textbf{right}). The mean AUROC is always reported on the left y-axis. The right y-axes show the cardinality of mined positive label names ($|\ypos|$) for different $M, C$. For $K=|\yneg|$ (left plot), all non-overlapping labels from the corpus are used (\cref{eq:negative_all}). Colors and markers per method are shared for all figures. Best viewed in color.}
    \label{fig:ablations_all}
    \label{fig:all_ablations_mine}
\end{figure*}

\textbf{Leveraging $\gt$ to mine positive labels from $\corpus$.}
In \Cref{tab:results_controlled}, to assess whether the higher cardinality of $\ypos$ alone (compared to $\gt$) is the mere reason for superior performance, we compare i) $|\ypos| \approx |\gt| (1000)$ for ImageNet by setting $C$ to $1.2K$ for ClusterMine, and ii) $|\ypos|=1500=|\gt \cup \yaug|$. To mine additional positive labels $\yaug$ using $\gt$, we augment $\gt$ with the 500 most similar labels from the corpus using top cosine similarity, and compare with MCM and using \cref{eq:posneg_score} as an OOD score. Results are reported in \cref{tab:results_controlled}. Intriguingly, increasing $|\ypos|$ alone is \textbf{not} improving performance (\cref{tab:results_controlled}) and using the $\gt$ positives as anchor does not increase OOD detection performance.

\subsection{Comparison with recent state-of-the-art}
While not an apples-to-apples comparison, in \cref{tab:new_sota_vitb}, we report results with ViT-B using ClusterMine with AdaNeg \cite{zhang2024ada_neg} (adapts negatives based on incoming OOD samples) and SynOOD \cite{li2025synood_iccv}, which requires synthetic sample generation and training. Still, \clustermine outperforms existing state-of-the-art methods on near-OOD benchmarks without a) access to $\gt$, b) synthetic sample generation, c) inference-based adaptations based on available OOD data.

\subsection{Ablation studies}

\noindent\textbf{Is negative label mining necessary?}
Jiang et al. \cite{jiang2024negative} suggest that ``the performance of the OOD detector is enhanced by incorporating more negative labels''. Their theoretical finding is in contrast with their negative mining strategy. In \Cref{fig:ablations_all} (left), we show the impact on AUROC from pruning negative labels from the text corpus for ViT-B \cite{radford2021clip} and the best-performing ViT-H dfn5b \cite{fang2023data_dfn}. AUROC scores deteriorate from negative label pruning.

\noindent\textbf{Hyperparameter sensitivity.} Here, we vary the number of minimum samples per class $M$ for \posmine in \Cref{fig:ablations_all} (center) and the number of clusters $C$ for \clustermine (right). As shown in \Cref{fig:ablations_all}, a critical aspect of ClusterMine is its \emph{insensitivity to C} due to the cluster voting step. Other recent methods, such as NegLabel, AdaNeg, and CLIPScope \cite{fu2024clipscope} require careful hyperparameter selection, which necessitates an OOD validation set and is challenging to determine for an arbitrary ID and text corpus.  

\section{Future work and conclusion}
We believe our work makes a significant step towards \emph{truly unsupervised OOD} using vision-language models by highlighting the importance of dynamically extracting suitable $\ypos$. Inarguably, a domain-relevant corpus is easier to satisfy than the exact knowledge of a fixed $\gt$. When no corpus exists, future work could explore the use of a large language model with a human in the loop to generate a domain-specific corpus for other application domains. This will reduce dependence on pre-existing text corpora.

In this work, we presented a general OOD detection label mining framework using CLIP. We demonstrate that vision language models are able to extract ID-related concepts from a text corpus without relying on $\gt$. We introduced \clustermine, which is robust to both covariate shifts and variations in the text corpus. \clustermine achieves state-of-the-art OOD detection scores across a wide range of large-scale OOD benchmarks, all while requiring no fine-tuning in the learned weights of CLIP models.

\bibliographystyle{ieeenat_fullname}
\bibliography{egbib}

\clearpage
\appendix
\section{Appendix}
\section{Cluster-based metrics for ClusterMine.}
\begin{figure*}[tb]
\centering\includegraphics[width=0.99\linewidth]{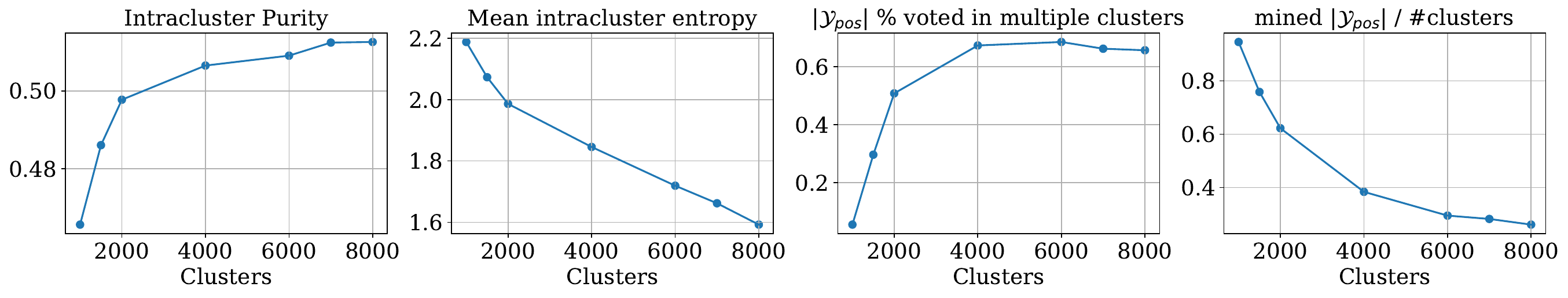}
\caption{\textbf{ClustMine cluster analysis for various cluster sizes.} From left to right: intracluster purity measures the percentage of samples within a cluster that have the most frequent (majority) label (\textit{left}), intracluster entropy is computed within samples in the same cluster (\textit{center left}), and we compute the percentage of mined $|\ypos|$ that appears across multiple clusters as $C$ increases (\textit{center right}), and finally we compute the ratio of the mined $|\ypos|$ related to the chosen number of clusters $C$ (\textit{right}).}
\label{fig:pure_all}
\end{figure*}

To further investigate the quality of the clusters, we compute in \Cref{fig:pure_all}: (i) the intercluster \textbf{purity} (percent of samples withing a cluster that share the majority label), (ii) the intercluster \textbf{entropy} w.r.t. mined labels $\ypos$, and (iii) how often (percentage) $\ypos$ appear across multiple clusters (redundancy ratio), (iv) the ratio of mined $|\ypos|/C$. We find that clusters typically exhibit high purity relative to the number of clusters ($\geq 50\%$). This further supports our assumption that feature-space neighbors are label/cluster-consistent. Interestingly, the redundancy ratio increases as $C$ increases, confirming that ClusterMine maintains semantic consistency while being robust to the overestimation of $C$. This analysis highlights the benefits of choosing ClusterMine over PosMine or NegLabel, where it is challenging to determine their respective hyperparameters in advance.

\textbf{Heuristic for picking C.} The redundancy ratio and the ratio of mined $|\ypos|/C$ could be used as a guideline to pick $C$ using the elbow approach. While increasing $C$, these ratios tend to saturate and can serve as informative label-free heuristics in new application domains.

\begin{table}[h]
\resizebox{1\columnwidth}{!}{%
\begin{tabular}{lcc|cc|ccc}
\toprule
& \multicolumn{2}{c}{No duplicates} & \multicolumn{2}{c}{Duplicates} & \multicolumn{2}{c}{Duplicates } \\
& \multicolumn{2}{c}{1 lemma} & \multicolumn{2}{c}{all lemmas} & \multicolumn{2}{c}{1 lemma} \\
  & AUROC & FPR95 & AUROC & FPR95 & AUROC & FPR95 \\
\midrule
NINCO & 92.87 & 30.30 & 92.89 & 29.86 & 92.80 & 30.18 \\
IN-O & 93.57 & 29.40 & 93.53 & 28.60 & 93.38 & 30.45 \\
OpenImage-O & 96.93 & 15.91 & 97.06 & 14.90 & 96.85 & 16.09 \\
iNat & 99.00 & 4.77 & 99.05 & 4.14 & 98.92 & 4.98 \\
IN-OOD & 91.53 & 38.26 & 91.14 & 38.35 & 91.28 & 38.77 \\
Textures43 & 93.45 & 32.89 & 94.10 & 27.61 & 93.99 & 29.82 \\
\midrule
Mean & 94.56 & 25.25 & 94.63 & 23.91 & 94.54 & 25.05 \\
\bottomrule
\end{tabular}
}
\caption{We report the impact of text-based preprocessing in WordNet (nouns and adjectives) using ClusterMine with CLIP ViT-H \cite{fang2023data_dfn}.}
\label{tab:text_prepro}
\end{table}

\textbf{Text pre-processing.} Homographs/duplicates words (e.g. bank) are deduplicated from the corpus, and we used only one lemma per Synset (no duplicates, one lemma in \cref{tab:text_prepro}). In WordNet, a SynSet represents a group of cognitive synonyms that convey a shared concept or meaning. Nonetheless, text pre-processing had a minuscule impact on the reported results using ClusterMine as shown in \cref{tab:text_prepro}. 

\section{Results using additional OOD datasets.} 
\cref{tab:extra_ood_data} reported the AUROC on four additional OOD datasets. The Places dataset has the highest semantic overlap with the ID ($\approx 60\%$), while NINCOv2 has a near-zero semantic overlap, as it is a manually picked collection from existing OOD datasets. In contrast to prior works, we use Places as a bad benchmark to showcase how OOD detectors can reject samples that are more likely to be ID. We observe that MCM is the best-performing approach on Places and the worst-performing approach on NINCOv2, respectively. These two benchmarks could be utilized as OOD validation sets in future work.

\begin{table}[hb]
\centering
\caption{\textbf{Semantic OOD detection detection AUROC on additional OOD datasets.}}
\resizebox{\linewidth}{!}{%
\begin{tabular}{lccccc} 
\toprule
Method  & Places & Texture & SUN & SSB & NINCOv2    \\ 
\midrule
MCM  & \textbf{92.32} & 90.40 & 94.37 & 79.69 & 92.50  \\
PLP & 92.15 & \textbf{93.08} & 94.01 & 81.42 & 94.72\\
NegLabel  & 90.08 & 83.13 & 93.70  & 82.80 & 93.07 \\
NegLabel$^*$  & 90.39 & 88.29 & 94.41 & 85.13 & 94.53  \\
PosMine  & 92.08 & 92.45 & \textbf{95.51} & 85.44 & 95.58  \\
ClusterMine  & 91.41 & 91.80 & 94.70 & \textbf{86.04} & \textbf{95.86} \\
\bottomrule
\end{tabular}}\label{tab:extra_ood_data}
\end{table}

\noindent\textbf{Near-OOD and far-OOD detection.}
Recent works \cite{ahmed2020detecting,zhang2023openood15} make a distinction between near-OOD and far-OOD based on image semantics or empirical difficulty. SSB, IN-OOD, and NINCO are considered near-OOD, while iNat, Textures, and OpenImage-O are considered far-OOD. Under this prism, \clustermine is the current state-of-the-art method on near-OOD detection. %

\begin{figure*}
\centering\includegraphics[width=0.9\linewidth]{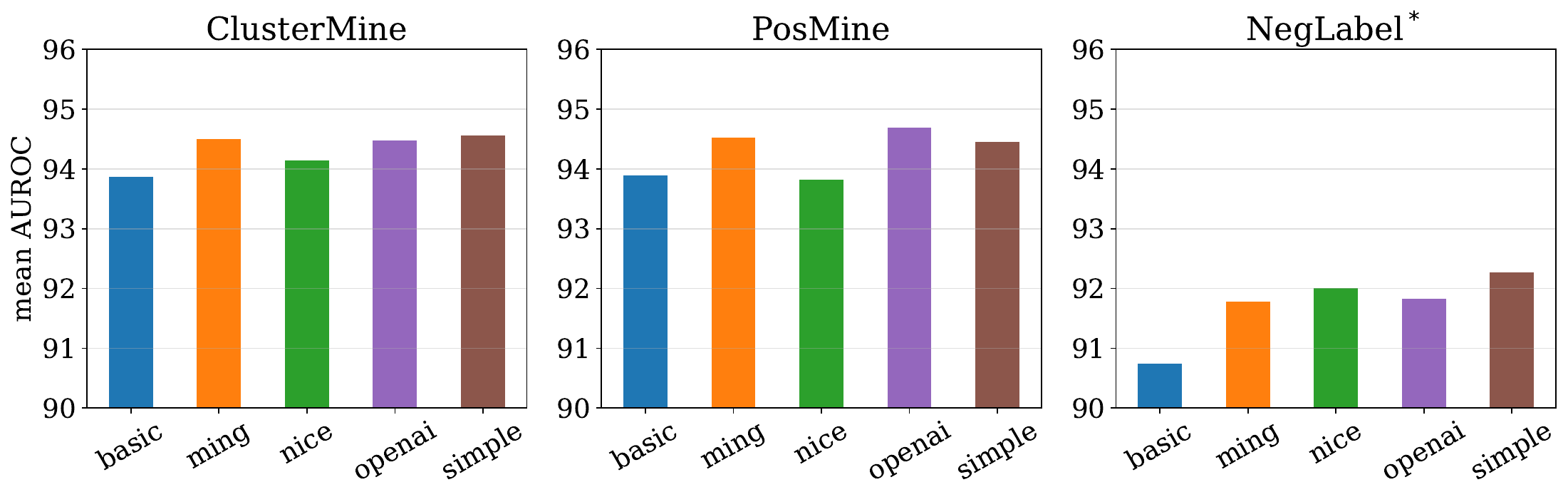}
    \caption{\textbf{Ablation study on context text prompts using CLIP VIT-H dfn5b \cite{fang2023data_dfn}}.} 
    \label{fig:ablation_prompt}
\end{figure*}

\section{Context prompt sensitivity.}
In \cref{fig:ablation_prompt}, we present the sensitivity of the label mining approaches to the different sets of context prompts. Basic refers to ``An image of a \{label\}'', while OpenAI refers to the set of 80 prompts as used by Radford et al. \cite{radford2021clip}. Ming refers to a subset of 5 prompts taken from the OpenAI set as used in \cite{ming2022mcm, adaloglou2023adapting}. Nice refers to ``A nice \{label\}'' used by Jiang et al. \cite{jiang2024negative}.
    Simple is a subset of 7 out of the 80 initial prompts, namely ``itap of a \{label\}'', ``a bad photo of the \{label\}'', ``an origami \{label\}'', ``a photo of the large \{label\}'', ``a \{label\} in a video game'', ``art of the \{label\}'', ``a photo of the small \{label\}'', based on follow-up analysis of Radford et al. \cite{radford2021clip} \url{https://github.com/openai/CLIP/blob/main/notebooks/Prompt_Engineering_for_ImageNet.ipynb}. We adopt the simple prompts for all the reported results in the main text.

\section{Additional results using negative label mining.}

\begin{figure}
\centering\includegraphics[width=0.99\linewidth]{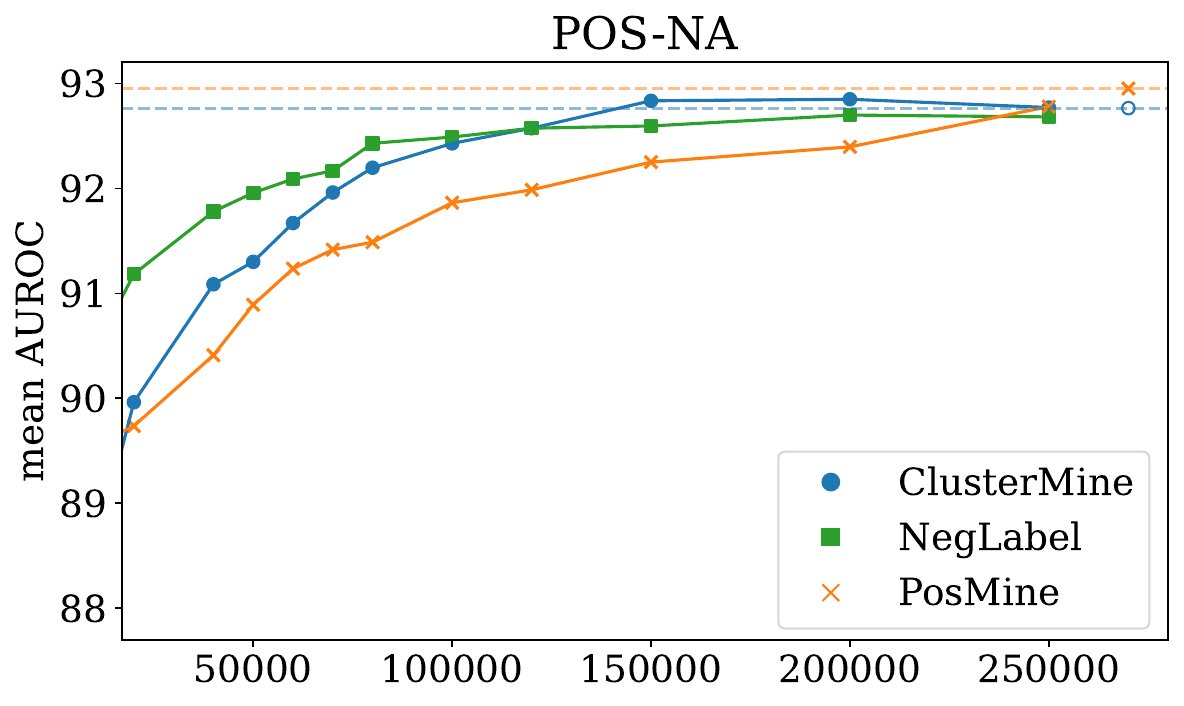}
    \caption{\textbf{The impact of negative label mining on the POS-NA corpus.}}
    \label{fig:neglabel_posna}
\end{figure}

\textbf{Negative label mining on POS.} \cref{fig:neglabel_posna} shows that negative label mining is not improving performance even for a large-scale text corpus such as POS. We included all available nouns and adjectives. 

\begin{figure}
\centering\includegraphics[width=0.99\linewidth]{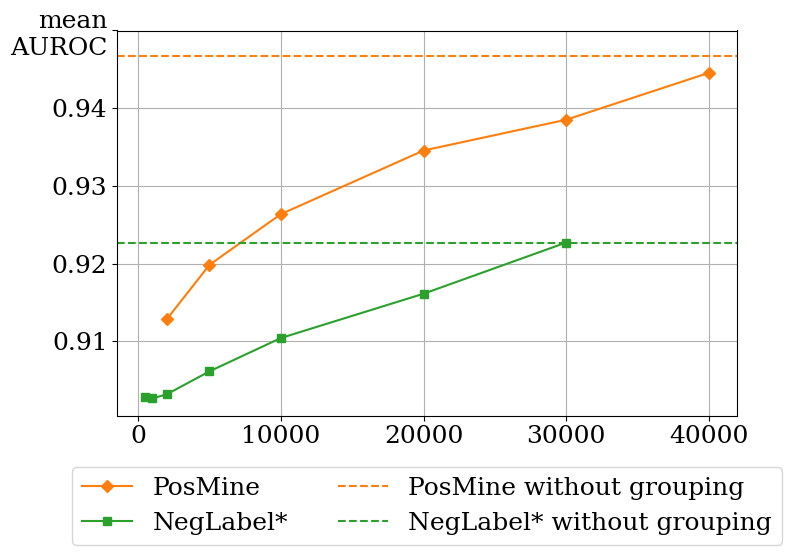}
    \caption{\textbf{Ablation study on the negative grouping strategy using CLIP VIT-H dfn5b \cite{fang2023data_dfn}}. The x-axis represents group size (i.e. number of label names per group). NegLabel$^*$ uses 40K
negative mined labels as in the main manuscript.}
    \label{fig:grouping}
\end{figure}

\textbf{Negative grouping strategy.} In \cref{fig:grouping}, we show that the proposed negative grouping strategy by NegLabel \cite{jiang2024negative} deteriorates the OOD detection AUROC. Thus we did not include it in ou experiments.

\begin{figure*}
\centering\includegraphics[width=0.99\linewidth]{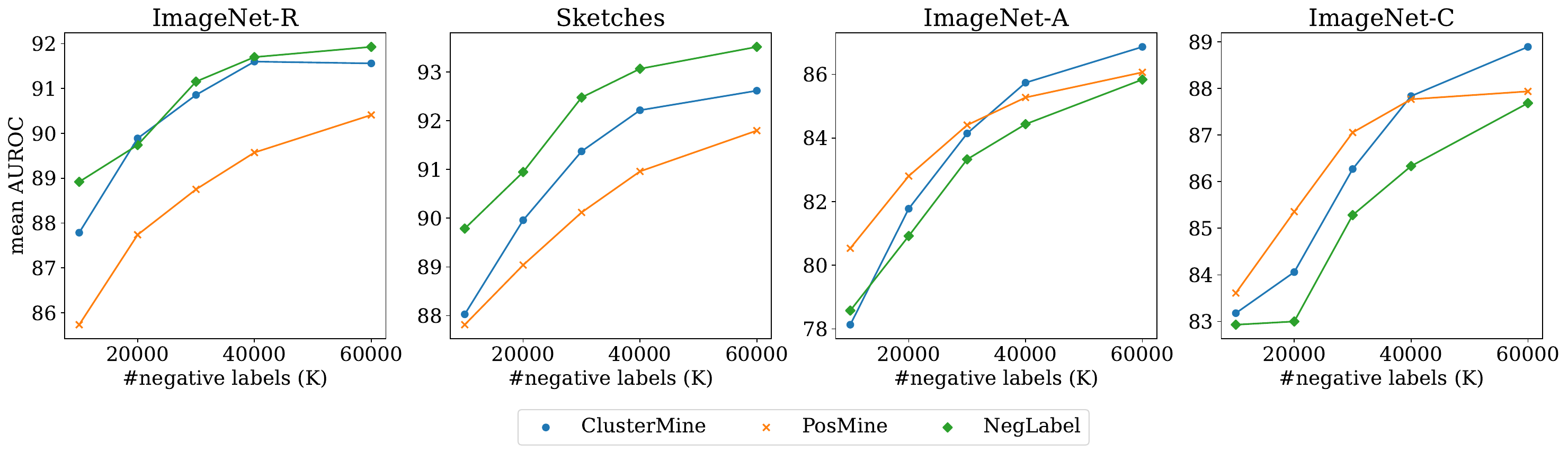}
\caption{\textbf{OOD detection robustness to covariate ID shifts.} The superior performance of NegLabel on ImageNet-R and Sketches is not attributed to the negative label mining but rather to the a priori knowledge of $\gt$, unlike \posmine and \clustermine.}
\label{fig:negmine_robust}
\end{figure*}

\textbf{Negative mining on various covariate ID sets.} In \cref{fig:negmine_robust}, we demonstrate that negative label mining is not the reason of superior performance on the datasets with stylistic perturbations, namely ImageNet-R and Sketches. Hence, the outperformance of NegLabel is primarily attributed to the \textit{a priori} knowledge of $\gt$.

\begin{table*}  
\centering
\label{tab:plp_analysis}
\caption{\textbf{Applying pseudo-label probing (PLP) using the derived $\ypos$ instead of $\gt$ from PosMine and ClusterMine, similar to \cite{adaloglou2023adapting}.}}
\begin{tabular}{lcccccccc} 
\toprule
\multirow{2}{*}{{Method}} & $\gt$ & \multirow{2}{*}{{NINCO}} & \multirow{2}{*}{{IN-O}}   & \multirow{2}{*}{{OpenImage-O}} & \multirow{2}{*}{{iNat}} & \multirow{2}{*}{{IN-OOD}} &   \multirow{2}{*}{{Textures43}}  & Average     \\ 
& {/$\corpus$} &&&&&&& AUROC / FPR95  \\ 
\midrule
 PLP \cite{adaloglou2023adapting} & \cmark/\xmark&  91.80 & 93.30 & \textbf{97.87} &  {98.94} & 90.01 & 94.79 & 94.45 / 27.29 \\
  \midrule

PosMine         & \colorbox{lightgray}{\xmark /\cmark} & 92.56 & 93.13 & 97.04 & 98.83 & 91.36 & 93.81 & 94.46 / 26.41 \\
PosMine + PLP & \colorbox{lightgray}{\xmark /\cmark} & 91.72 & 92.79 & 97.43 & 98.68 & 88.56 & 94.38 & 93.93 / 27.79 \\
ClusterMine     &  \colorbox{lightgray}{\xmark /\cmark} & \textbf{92.87} & \textbf{93.57} & 96.93 & \textbf{99.00} & \textbf{91.53} & 93.45 & \textbf{94.56} / \textbf{25.26} \\
ClusterMine + PLP & \colorbox{lightgray}{\xmark /\cmark} & 92.25 & 93.07 & 97.46 & 98.79 & 88.89 & 94.71 & 94.20 / 27.53 \\

\bottomrule
\end{tabular}
\end{table*}

\section{Combining pseudo-label-probing (PLP) with positive label mining.}
Finally, we explore whether pseudo-label probing (PLP) can be combined with positive label mining in \cref{tab:plp_analysis}. We found no significant performance gain by combining PLP with ClusterMine and PosMine.

\begin{table*}
\centering
\caption{\textbf{Quantifying robustness to in-distribution shifts using CLIP ViT-H dfn5b \cite{fang2023data_dfn}.} We report AUROC (\%,$\uparrow$) per OOD detection benchmark as well as mean AUROC and mean FPR95 (\%,$\downarrow$). Results with MCM \cite{ming2022mcm} and NegLabel \cite{jiang2024negative}, where NegLabel$^*$ uses 40K negative mined labels. The highlighted \colorbox{lightgray}{light gray} values highlight the discussion point raised in the main manuscript.}
\label{tab:indist-shifts}
\begin{tabular}{@{}llcccccccccc@{}}
\toprule
\multirow{2}{*}{{ID Dataset}} & \multirow{2}{*}{{Method}} & \multirow{2}{*}{{NINCO}} & \multirow{2}{*}{{IN-O}}   & \multirow{2}{*}{{OpenImage-O}} & \multirow{2}{*}{{iNat}} & \multirow{2}{*}{{IN-OOD}} &   \multirow{2}{*}{{Textures43}}  & Average & Average     \\ 
  &  &  &    & &  &  & & AUROC & FPR95  \\
\hline
                                
  & MCM & 84.88 & 87.73 & 94.18 & 94.01 & 85.77 & 88.22 & 89.13 & 55.74 \\
  & NegLabel$^*$ & 91.21 & 91.11 & 96.31 & 98.85 & 89.50 & 91.41 & \colorbox{lightgray}{\textbf{93.06}} & \colorbox{lightgray}{40.29} \\
  & PosMine & 89.60 & 90.44 & 95.68 & 98.24 & 88.14 & 91.25 & \colorbox{lightgray}{92.22} & \colorbox{lightgray}{32.92} \\
 \multirow{-4}{*}{\makecell{ImageNet-\\Sketches}}  & ClusterMine & 90.35 & 91.14 & 95.60 & 98.52 & 88.43 & 90.76 & \colorbox{lightgray}{92.47} & \colorbox{lightgray}{\textbf{30.94}} \\
 \midrule
  & MCM & 77.67 & 81.64 & 91.08 & 90.74 & 78.77 & 82.11 & 83.66 & 68.28 \\
  & NegLabel$^*$ & 89.30 & 89.48 & 95.44 & 98.59 & 87.71 & 89.68 & \colorbox{lightgray}{\textbf{91.70}} & \colorbox{lightgray}{41.42} \\
  & PosMine & 87.74 & 88.69 & 94.91 & 98.00 & 86.01 & 89.58 & \colorbox{lightgray}{90.82} & \colorbox{lightgray}{\textbf{35.92}} \\
 \multirow{-4}{*}{ImageNet-R} & ClusterMine & 88.34 & 89.13 & 94.50 & 98.17 & 85.91 & 88.61 & \colorbox{lightgray}{90.78} & \colorbox{lightgray}{37.50} \\
 \midrule
  & MCM & 62.95 & 68.79 & 84.49 & 83.75 & 64.15 & 69.05 & 72.20 & 76.80 \\
  & NegLabel$^*$ & 79.70 & 80.82 & 90.74 & 96.64 & 78.14 & 80.56 & 84.43 & 56.32 \\
  & PosMine & 82.13 & 83.52 & 92.19 & 96.71 & 79.92 & 84.64 & 86.52 & 44.44 \\
  \multirow{-4}{*}{ImageNet-A}  & ClusterMine & 83.53 & 84.45 & 91.88 & 97.17 & 80.04 & 83.46 & \textbf{86.76} & \textbf{42.57} \\
  \midrule
  & MCM & 84.72 & 87.79 & 94.79 & 94.66 & 85.66 & 88.21 & 89.31 & 48.97 \\
  & NegLabel$^*$ & 87.91 & 87.97 & 94.60 & 98.17 & 86.09 & 88.22 & 90.49 & 45.38 \\
  & PosMine & 90.44 & 91.10 & 95.94 & 98.31 & 89.00 & 91.86 & \textbf{92.78} & 32.15 \\
 \multirow{-4}{*}{ImageNetV2}  & ClusterMine & 90.70 & 91.39 & 95.65 & 98.49 & 88.85 & 91.10 & 92.70 & \textbf{31.87} \\ \midrule
  & MCM & 70.01 & 73.90 & 84.58 & 83.76 & 70.76 & 74.43 & 76.24 & 89.99 \\
  & NegLabel$^*$ & 82.68 & 83.18 & 91.54 & 96.49 & 80.95 & 83.17 & 86.33 & 54.87 \\
  & PosMine & 84.86 & 85.61 & 92.60 & 96.47 & 82.80 & 86.58 & 88.15 & 52.10 \\
\multirow{-4}{*}{ImageNet-C}  & ClusterMine & 86.40 & 87.02 & 92.81 & 97.11 & 83.75 & 86.54 & \textbf{88.94} & \textbf{49.21} \\ \midrule
  & MCM & 88.78 & 91.30 & 96.64 & 96.62 & 89.65 & 91.75 & 92.46 & 35.22 \\
  & NegLabel$^*$ & 90.26 & 90.10 & 95.79 & 98.64 & 88.40 & 90.44 & 92.27 & 40.43 \\
  & PosMine & 92.56 & 93.13 & 97.04 & 98.83 & 91.36 & 93.81 & 94.46 & 24.78 \\
    \multirow{-4}{*}{\textit{\makecell{ImageNet\\ ID test set}}} & ClusterMine & 92.87 & 93.57 & 96.93 & 99.00 & 91.53 & 93.45 & \textbf{94.56} & \textbf{24.23} \\
\bottomrule
\end{tabular}%
\end{table*}

\section{Robustness to covariate ID shifts.}
\Cref{tab:indist-shifts} shows all the individual robustness scores for the sensitivity to ID shifts. In the main text we report the mean AUROC and FPR95.

\end{document}